\pdfoutput=1
\documentclass[11pt]{article}

\usepackage[final]{acl}

\usepackage{times}
\usepackage{latexsym}
\usepackage{amsmath}
\usepackage{booktabs}
\usepackage{tcolorbox}
\usepackage{listings}
\usepackage{float}
\usepackage[T1]{fontenc}
\usepackage[utf8]{inputenc}
\usepackage{subcaption}
\usepackage{algorithm}
\usepackage{algpseudocode}
\usepackage{hyperref}

\usepackage{microtype}
\usepackage{inconsolata}
\usepackage{graphicx}

\title{DiaSynth: Synthetic Dialogue Generation Framework for Low Resource Dialogue Applications}

\author{
  \textbf{Sathya Krishnan Suresh\textsuperscript{1}},
  \textbf{Wu Mengjun\textsuperscript{1}},
  \textbf{Tushar Pranav\textsuperscript{2}},
  \textbf{Eng Siong Chng\textsuperscript{1}}
\\
  \textsuperscript{1}Nanyang Technological University 
  \textsuperscript{2}Singapore Institute of Technology
\\
  \small{\textbf{Correspondence:} \href{mailto:sathyakr001@e.ntu.edu.sg}{sathyakr001@e.ntu.edu.sg}}
}

\begin{document}
\maketitle
\begin{abstract}
The scarcity of domain-specific dialogue datasets limits the development of dialogue systems across 
applications. Existing research is constrained by general or niche datasets that lack sufficient scale 
for training dialogue systems. To address this gap, we introduce \textbf{DiaSynth} - a synthetic 
dialogue generation framework capable of generating high-quality, contextually rich dialogues across 
a wide range of domains. Unlike existing frameworks, DiaSynth uses Large Language Models (LLMs) and 
Chain of Thought (CoT) reasoning to generate dynamic, domain-specific dialogues with simulated personas 
and diverse conversational features. We perform our experiments by generating synthetic data using 
different LLMs and few-shot examples from DialogSum and SAMSum. The pretrained language models fine-tuned 
on the synthetic data outperform the base models by \textbf{16.47\%} on dialogue summarization, 
while the comparison between models fine-tuned on in-domain data and synthetic data shows that 
the synthetic data is able to capture \textbf{90.48\%} of the performance distribution of the 
in-domain data on dialogue summarization. The quality of the data generated also increases as we 
increase the size of LLM from 3B to 8B. These results validate DiaSynth's potential as a robust 
alternative to traditional data collection methods.\footnote{\url{https://github.com/ntuspeechlab/DiaSynth}} 
We open source the code and data generated for future research.
\end{abstract}

\section{Introduction}

\begin{figure*}[htbp]
    % \centering
    \includegraphics[width=\textwidth]{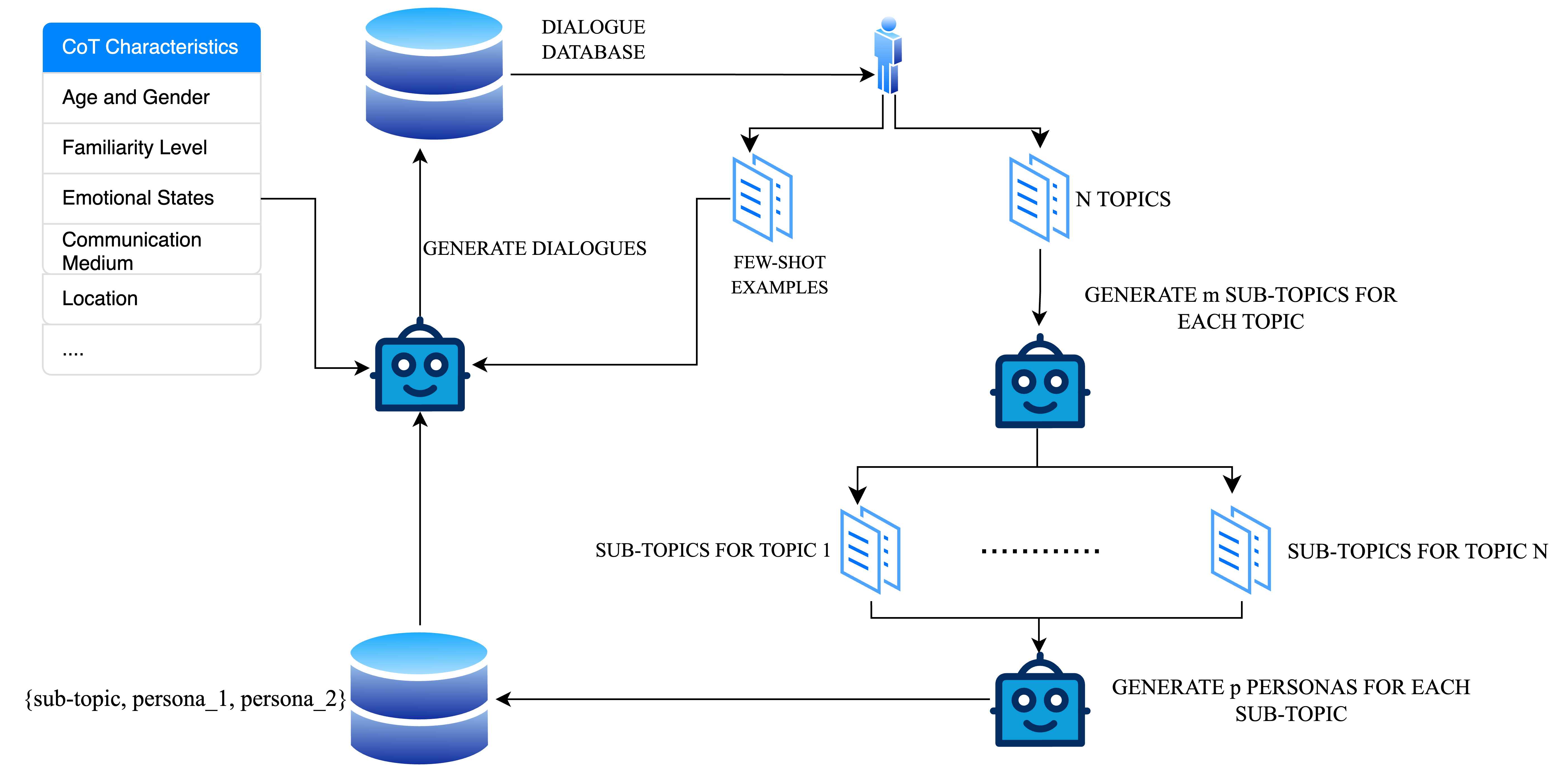}
    \caption{Overview of DiaSynth}
    \label{fig:dialogue_synthesis}
\end{figure*}

Dialogue systems are crucial in natural language processing due to applications like customer service chatbots and virtual assistants. Their effectiveness depends on large, high-quality, domain-specific datasets. The lack of large-scale, high-quality datasets across domains like academic discussions, healthcare, and everyday conversations poses a challenge. This scarcity limits the development of dialogue systems that generalize well across domains.

Prior work (\citet{feng-etal-2020-doc2dial}, \citet{zeng-etal-2020-meddialog}, \citet{budzianowski-etal-2018-multiwoz}) collects domain-specific dialogues but often lacks depth, scale, or domain diversity. On the one hand, the conversations in a domain specific dataset are superficial and do not go deep into the domain. On the other hand, niche domain dialogue datasets, while contextually rich, often suffer from limited scale. This imbalance hinders dialogue system development in underrepresented domains, where data collection is costly and complex.

To address these problems, we introduce \textbf{DiaSynth}, a synthetic dialogue generation framework that produces contextually rich and realistic dialogues tailored to specific domains. DiaSynth, using a Large Language Model (LLM), generates high-quality conversations by simulating personas and conversation characteristics like tone and formality. With LLMs and Chain of Thoughts (CoT) \citet{cot}, DiaSynth generates dialogues that mimic real-world conversations for a wide range of domains. CoT plays a crucial part by building different environments for different personas (refer to Appendix \ref{app:cot_examples} for examples) which influence varied conversations. This approach addresses data scarcity and offers a scalable, cost-effective alternative to traditional methods.

To validate the effectiveness of DiaSynth, we evaluated the framework on two criteria: the quality of the generated data and the usability of this data for downstream tasks. The results for the quality criterion showed that the data quality improved with the scale of the model. For usability, we tested DiaSynth on dialogue summarization. Models fine-tuned on DiaSynth data outperformed base versions by \textbf{16.47\%} on average. Additionally, DiaSynth's synthetic data captures \textbf{90.48\%} of in-domain performance, highlighting its potential as a strong alternative when domain-specific data is unavailable.

The remainder of this paper is organized as follows: Section \ref{sec:related_work} reviews related work on dialogue datasets and synthetic data. Section \ref{sec:diasynth} details the DiaSynth framework and methodology. Section \ref{sec:experiments} describes our experimental setup and evaluation. Section \ref{sec:results} compares the performance of models fine-tuned on DiaSynth data to in-domain data.  Section \ref{sec:conclusion} concludes with a summary of our findings and Section \ref{sec:limitations} discusses the limitations of DiaSynth and potential future directions for this research.

\section{Related Work}
\label{sec:related_work}

\subsection{Personality in Synthetic Data Generation}

In recent years, there has been a significant increase in research focused on synthetic dialogue generation, largely driven by advancements in Large Language Models (LLMs). To generate \textbf{realistic and diverse} synthetic data, researchers have incorporated personalities, profiles, and character information when prompting LLMs to generate dialogues \citet{han2024psydial}. By enhancing dialogue realism through the simulation of various personality profiles, utilizing the Big Five personality model, and employing structured prompts, this approach has improved task performance in models fine-tuned on these generated dialogues compared to those trained on general chit-chat datasets.

Moreover, integrating personas into synthetic data generation prompts \citet{chan2024scaling} has demonstrated that models fine-tuned on personalized synthetic data outperform some LLMs of much larger scales. The inclusion of personas in prompts provides diversity in difficulty levels and ranges within the synthetic data, enabling the models to handle situations of varying complexity.

Our approach involves persona extraction after generating subtopics related to the general topics. This enables the generated dialogues to be more in-depth and specific to those subtopics, enhancing both the scale and quality of domain-specific dialogue generation.

\subsection{Prompting Task-Oriented Dialogue Generation}

\begin{table*}[htbp]
    \centering
    \resizebox{\textwidth}{!}{%
    \begin{tabular}{cccccc}
    \toprule
    \textbf{LLM} & \textbf{Few-shot examples} & \textbf{Number of Samples}  & \textbf{Avg. number of turns} & \textbf{Avg. number of tokens per turn} & \textbf{Diversity (ROUGE-L)} \\
    \midrule

    Phi-3 & DialogSum & 1215 & 9.13 & 20.38 & 0.27 \\
    InternLM-2.5 & DialogSum & 1035 & 9.23 & 27.98 & 0.30 \\  
    LLaMA-3 & DialogSum & 1154 & 6.86 & 31.99 & 0.29 \\
    GPT-4o & DialogSum & 1375 & 15.16 & 15.96 & 0.29 \\
    Phi-3  & SAMSum & 1410 & 13.98 & 13.94 & 0.27 \\
    InternLM-2.5 & SAMSum & 1135 & 13.96 & 19.07 & 0.29 \\ 
    LLaMA-3 & SAMSum & 1195 & 10.54 & 20.41 & 0.29 \\
    GPT-4o & SAMSum & 1380 & 15.43 & 13.53 & 0.28 \\    
    \bottomrule
    \end{tabular}%
    }
    \caption{Data Statistics}
    \label{tab:data_stats}
\end{table*}

Prompt-based techniques have also emerged as powerful methods for generating high-quality synthetic dialogues, particularly for task-oriented dialogue systems. \citet{cheng2024generating} explore the generation of synthetic dialogues from structured prompts, focusing on enhancing task-oriented dialogue systems. Their work demonstrates that prompt engineering can produce dialogues that are contextually appropriate and improve system performance by aligning synthetic data more closely with real-world requirements.

To achieve a \textbf{higher quantity, diversity, and creativity} in human-written instruction data, \citet{wang2022self} propose inputting prompts to LLMs to generate instructions based on a small set of seed human-written instructions. This approach aligns the expanded training data more closely with desired task objectives and allows for iterative improvements, producing more nuanced and effective dialogues that meet specific task demands.

Similarly, our study expands topics into subtopics, ensuring that the generated dialogues provide more in-depth and high-quality conversations. By doing so, we aim to produce synthetic data that not only covers a broader range of scenarios but also delves deeper into each topic, thereby enhancing the overall effectiveness of the dialogue systems trained on this data.

\subsection{Existing Task-Oriented Dialogue Datasets}

In addition to prompt-based synthetic data generation, various large-scale dialogue datasets have been instrumental in advancing task-oriented dialogue systems. Among these, the MultiWOZ dataset \citet{budzianowski-etal-2018-multiwoz} is a prominent resource, providing richly annotated dialogues across multiple domains. MultiWOZ has enabled researchers to train models capable of handling complex, multi-turn interactions across diverse tasks. The nature of MultiWOZ's annotations has made it a benchmark for evaluating the performance of dialogue systems, though it is often complemented by synthetic data to introduce further diversity and variation in dialogue scenarios.

Similarly, Doc2Dial \citet{feng-etal-2020-doc2dial} is another widely used dataset designed specifically for document-grounded dialogue systems. Doc2Dial includes conversations grounded in structured documents, focusing on providing users with accurate and relevant information based on their inquiries. This dataset has been instrumental in improving the ability of dialogue systems to retrieve and generate accurate responses when interacting with complex information sources. However, much like MultiWOZ, Doc2Dial's scope is limited to the predefined topics and domains covered within the dataset, which can restrict model generalizability to new or unseen situations.

To overcome the domain-specific limitations of datasets, our study adopts a synthetic data generation approach that expands existing topics into subtopics, thus providing a broader and deeper pool of conversational data. By incorporating both task-oriented prompts and personas, our generated dialogues aim to complement these datasets by offering more personalized and contextually rich conversations, thereby enhancing the robustness and versatility of dialogue systems

\section{DiaSynth}
\label{sec:diasynth}

\begin{table*}[htbp]
    \centering
    \resizebox{\textwidth}{!}{
\begin{tabular}{lcccccccccc}
\toprule
 & \textbf{coherent} & \textbf{error recovery} & \textbf{consistent} & \textbf{diverse} & \textbf{depth} & \textbf{likeable} & \textbf{understand} & \textbf{flexible} & \textbf{informative} & \textbf{inquisitive}\\
\midrule
\multicolumn{11}{c}{\textbf{DIALOGUESUM}} \\
\midrule

\textbf{Phi-3} & 0.9536 & 0.9440 & 0.9540 & 0.9534 & 0.9521 & -0.0005 & 0.9353 & \textbf{-3.96E-05} & 0.0009 & -0.0033 \\

\textbf{InternLM-2.5} & 0.8439 & 0.8313 & 0.8359 & 0.8353 & 0.8352 & 0.0048 & 0.8278 & -0.0046 & 0.0042 & 0.0069 \\

\textbf{LLaMA-3} & \textbf{0.9684} & \textbf{0.9522} & \textbf{0.9570} & \textbf{0.9596} & \textbf{0.9592} & 0.0032 & \textbf{0.9453} & -0.0063 & 0.0063 & 0.0105 \\

\textbf{GPT-4o} & 0.9525 & 0.9407 & 0.9417 & 0.9423 & 0.9425 & \textbf{0.0121} & 0.9368 & -0.0027 & \textbf{0.0085} & \textbf{0.0144} \\

\midrule

\multicolumn{11}{c}{\textbf{SAMSUM}} \\
\midrule
\textbf{Phi-3} & 0.9161 & 0.9088 & 0.9199 & 0.9161 & 0.9130 & -0.0004 & 0.9014 & -0.0024 & 0.0034 & -0.0040 \\
\textbf{InternLM-2.5} & 0.8746 & 0.8647 & 0.8734 & 0.8655 & 0.8661 & 0.0033 & 0.8582 & -0.0019 & 0.0106 & 0.0028 \\
\textbf{LLaMA-3} & 0.9829 & 0.9677 & 0.9757 & 0.9712 & 0.9731 & 0.0003 & 0.9593 & -0.0048 & 0.0100 & 0.0029 \\

\textbf{GPT-4o} & \textbf{0.9939} & \textbf{0.9876} & \textbf{0.9878} & \textbf{0.9836} & \textbf{0.9824} & \textbf{0.0083} & \textbf{0.9788} & \textbf{0.0004} & \textbf{0.0141} & \textbf{0.006} \\

\bottomrule
\end{tabular}
}
    \caption{FED scores}
    \label{tab:fed_scores}
\end{table*}

DiaSynth is a synthetic dialogue generation framework designed to address the scarcity of high-quality, large-scale, domain-specific dialogue datasets. DiaSynth uses an LLM and CoT reasoning to simulate diverse, nuanced dialogues.

DiaSynth takes a list of user-provided topics to generate dialogues. The users can optionally provide few-shot examples of the format in which they want the dialogue to be generated. Directly generating dialogues from user topics would be too superficial due to their lack of specificity. To overcome this lack of specificity, we generate $m$ sub topics for each of the $n$ topics given by the user. Generating dialogues from the subtopics will have specificity but the dialogues will lack variety. This is because every dialogue is influenced implicitly by the personas of the people involved in the dialogue and, other characteristics such as the location, emotion and more. To enhance variety and depth, we generate $p$ personas per subtopic and create dialogues for all persona-subtopic combinations. To further ground the dialogues in various settings and characteristics, we employ CoT reasoning during the generation process. DiaSynth employs CoT to reason about the settings and characteristics of a dialogue, which are listed in Appendix \ref{app:conv_characteristics}, ensuring that the dialogues are contextually rich and realistic. This three-stage pipeline not only guarantees the quality of the generated dialogues but also allows for exponential scalability, as illustrated programmatically in Appendix \ref{app:scalability}.

\subsection{Subtopic Generation}

Subtopic generation is a crucial step in DiaSynth's pipeline, since it enhances the specificity and 
depth of the dialogues that will be generated later. For each primary topic given by the user, 
DiaSynth generates multiple subtopics, effectively narrowing the focus of the conversation. This 
breakdown is necessary because the primary topics are often too general to generate contextually 
rich dialogues on their own. For instance, a topic like “healthcare” can be expanded into subtopics 
such as “doctor-patient consultations,” “mental health discussions,” and “medical diagnostics,” 
each of which offers a more focused context for dialogue generation. To achieve this, DiaSynth 
prompts an LLM to generate the user specified number of subtopics for each primary topic. We also 
run a similarity check between each of the subtopics and remove subtopics that are too similar to 
other subtopics using a threshold.  

\subsection{Persona Generation}

Personas of the individuals involved in a conversation are primary influencers in determining how a 
conversation pans out. Using random personas from persona datasets and prompting the LLM to simulate 
a dialogue between them about a random topic often leads to superficial dialogues that lack depth 
and contextual richness. To address this issue, DiaSynth generates a user-specified number of personas 
for each subtopic, ensuring that the personas are conditioned on the subtopic context. This 
conditioning prompts the LLM to create personas that are most likely to engage in a meaningful 
dialogue about the subtopic, such as a medical professional and a patient discussing "medical diagnostics" 
or a researcher and a student talking about "academic publishing." We also run a similarity check 
for the personas too. The conditioned persona generation is crucial because it ensures that future 
dialogues will not only be contextually rich but also exhibit a high level of depth. Each dialogue 
will be between two personas who have relevant expertise or perspectives on the given subtopic, 
allowing the conversation to explore nuances that would otherwise be missed in a generic dialogue 
setting. We present the impacts of sub-topics and personas in Appendix \ref{app:ablation_subtopics_personas}

\subsection{Dialogue Generation}

The final stage in DiaSynth's pipeline is the generation of dialogues, where all the 
components—subtopics, personas, and characteristics—converge to create contextually
rich and realistic conversations. This step uses an LLM as the backbone and CoT as the 
reasoning mechanism, allowing the model to simulate dialogues that incorporate various aspects 
of human interaction. DiaSynth generates dialogues by pairing all persona-subtopic combinations. The 
process also integrates predefined characteristics (Table \ref{tab:dialogue_characteristics}) like 
emotional state, formality, and familiarity to guide the flow and style. These characteristics are 
defined in the CoT prompt, guiding the LLM to generate realistic, contextually appropriate dialogues. 
The importance of CoT and the lack of it affects the quality of the dialogues, which is shown 
quantitatively in Appendix \ref{app:ablation_cot}.

\section{Experimental Setup}
\label{sec:experiments}

\begin{table*}[htbp]
\centering
\resizebox{\textwidth}{!}{
\begin{tabular}{lcccccccccc}
\toprule
 & \textbf{coherence} & \textbf{diversity} & \textbf{flexibility} & \textbf{understandability} & \textbf{inquisitiveness} & \textbf{consistency} & \textbf{informativeness} & \textbf{likeability} & \textbf{depth} & \textbf{error recovery} \\
\midrule
\multicolumn{11}{c}{\textbf{DIALOGUESUM}} \\
\midrule
\textbf{Phi-3} & \textbf{0.0286} & 0.0310 & \textbf{0.0218} & 0.0193 & 0.0363 & \textbf{0.0369} & 0.0172 & \textbf{0.0213} & 0.0117 & 0.0342 \\
\textbf{InternLM-2.5} & 0.0069 & 0.0196 & 0.0084 & 0.0061 & 0.0244 & 0.0137 & 0.0148 & 0.0050 & 0.0080 & 0.0197 \\
\textbf{LLaMA-3} & 0.0189 & \textbf{0.0430} & 0.0186 & \textbf{0.0220} & \textbf{0.0415} & 0.0321 & \textbf{0.0318} & 0.0110 & \textbf{0.0201} & \textbf{0.0440} \\

\textbf{GPT-4o} & 0.0039 & 0.0156 & 0.0059 & 0.0039 & 0.018 & 0.0080 & 0.0097 & 0.0026 & 0.0053 & 0.0135 \\

\midrule
\multicolumn{11}{c}{\textbf{SAMSUM}} \\
\midrule
\textbf{Phi-3} & \textbf{0.0325} & 0.0372 & 0.0260 & 0.0270 & 0.0395 & 0.0415 & 0.0201 & \textbf{0.0232} & 0.0126 & 0.0290 \\
\textbf{InternLM-2.5} & 0.0128 & 0.0408 & 0.0194 & 0.0174 & 0.0504 & 0.0306 & 0.0328 & 0.0142 & 0.0177 & 0.0407 \\
\textbf{LLaMA-3} & 0.0288 & \textbf{0.0655} & \textbf{0.0306} & \textbf{0.0365} & \textbf{0.0622} & \textbf{0.0612} & \textbf{0.0542} & 0.0168 & \textbf{0.0270} & \textbf{0.0558} \\

\textbf{GPT-4o} & 0.0055 & 0.0162 & 0.0094 & 0.0084 & 0.0186 & 0.0119 & 0.0129 & 0.0038 & 0.0079 & 0.0199 \\

\bottomrule
\end{tabular}
}
\caption{GPTScore}
\label{tab:gpt_scores}
\end{table*}

In this section, we detail the experimental setup used to evaluate the effectiveness of DiaSynth. 
Our evaluation focuses on two criteria - quality of the dialogues generated and usability of the 
dialogues generated for a downstream task. Quality of the dialogues is evaluated using metrics such as 
FED, GPTScore, and G-Eval. We evaluate the usability of DiaSynth-generated dialogues by using summarization 
as the downstream task.

\subsection{Quality of the dialogues}

To evaluate the quality of the dialogues, we employ the metrics that have been developed for evaluating the quality of text generated by LLMs. We use the following metrics:
\begin{itemize}
    \item \textbf{FED} \citet{mehri-eskenazi-2020-unsupervised} - FED evaluates dialogue quality by comparing the probabilities of appended positive and negative utterances.
    \item \textbf{GPTScore} \citet{fu-etal-2024-gptscore} - GPTScore assesses dialogue quality by asking an LLM to evaluate criteria like coherence and diversity, with scores based on the probability of affirmative responses.
    \item \textbf{G-Eval} \citet{liu-etal-2023-g} - G-Eval rates dialogue on a 1-3 scale across criteria, with final scores as a weighted average from the LLM’s probability distribution.
\end{itemize}

To validate the framework across models and also domains, we generate data using three open source LLMs, one closed source LLM and also use few shot examples from two different dialogue datasets. The open sourced LLMs are - \textbf{Phi-3} \citet{abdin2024phi3}, \textbf{InternLM-2.5} \citet{Cai2024InternLM2TR}, \textbf{LLaMA-3} \citet{dubey2024llama} and the closed source LLM used is \textbf{GPT-4o}. The 8-bit quantized versions of the open source LLMs were used for faster experimentation and generation. The two different dialogue datasets that were used as few-shot examples are DialogSum \citet{chen-etal-2021-dialogsum} and SAMSum \citet{gliwa-etal-2019-samsum}

\subsection{Downstream Task - Summarization}

To evaluate the usability of the dialogues generated by DiaSynth, we choose summarization as the 
downstream task. Summarization, a key application of dialogue systems, aims to generate concise, 
contextually relevant summaries. We use three established evaluation metrics—QAGS, BERTScore, and 
ROUGE-L—to assess the performance of summarization models fine-tuned on DiaSynth-generated data.

\begin{itemize}
    \item \textbf{QAGS} \citet{qags} - QAGS evaluates factual consistency by generating questions from the summary and comparing answers to those from the source dialogue.
    \item \textbf{BERTScore} \citet{Zhang*2020BERTScore:} - BERTScore measures semantic similarity between generated and reference summaries using contextual embeddings.
    \item \textbf{ROUGE-L} \citet{lin-2004-rouge} - ROUGE-L measures longest common subsequence (LCS) overlap between generated and reference summaries.
\end{itemize}

We fine-tune pretrained summarization models like DistilBART, BART \citet{lewis-etal-2020-bart}, 
T5 \citet{raffel2020exploring} and LED \citet{beltagy2020longformer}, on DiaSynth-generated dialogues 
and evaluate their performance using the above metrics. We evaluate the usability of DiaSynth in two 
key aspects: first, by assessing the performance improvement of models fine-tuned on DiaSynth-generated 
data compared to the pretrained models; and second, by measuring the extent to which DiaSynth-generated 
data reflects real-world data distribution by comparing the performance of models fine-tuned on 
DiaSynth data versus those fine-tuned on in-domain data. We also present the results of finetuning BART and 
T5 on synthetic data and in-domain data on response generation in Appendix \ref{app:response_generation}.

\section{Results}
\label{sec:results}

This section discusses the results of the data generated using DiaSynth 
(quality of the data and usability in downstream tasks) with different LLMs and varying few-shot 
examples. Specifically, we utilized Phi-3, InternLM-2.5, LLaMA-3 and GPT-4o as the LLM backbones, 
and the few-shot examples were sourced from DialogueSum and SAMSum datasets. These combinations allow 
us to evaluate the robustness and adaptability of DiaSynth across different models and few shot examples. 
In total, eight distinct datasets were generated using DiaSynth by pairing each LLM with the two sets of 
few-shot examples, resulting in all possible combinations. For each combination, DiaSynth was provided 
with the same 16 broad topics and tasked with generating 6 subtopics for each topic, followed by creating 
6 personas for each subtopic. The statistics of the datasets generated using DiaSynth, including the 
number of dialogues, average number of turns, and average number of tokens per turn, are summarized in 
Table \ref{tab:data_stats}. All the experiments were run on a single A100 GPU with the generation time 
ranging from 2 hours to 4 hours. Additional methodolgical details are presented in Appendix \ref{app:method_details}.

\subsection{Quality of the Dialogues}

The quality of the synthetic datasets produced by DiaSynth was evaluated using FED, GPTScore, and G-Eval metrics, as detailed in Tables \ref{tab:fed_scores}, \ref{tab:gpt_scores}, and \ref{tab:g_eval_scores}. The results reveal distinct variations in performance across different model and dataset configurations, reflecting the unique characteristics of each.

\subsubsection{Metric Scores}

\textbf{FED:} The FED scores in Table \ref{tab:fed_scores} show that LLaMA-3 and GPT-4o achieve almost a perfect score (+1) in most of the criteria, while Phi-3 and InternLM-2.5 also have decent performances. GPT-4o has a clear advantage when it comes to generating likeable dialogues while there is not much separation on other criteria. 

\noindent \textbf{GPTScore:} Results illustrated in \ref{tab:gpt_scores} are surprising in that GPT-4o is the worst performing model on GPTScore, which might require further research while LLaMA-3 clearly dominates the other models. 

\noindent \textbf{G-Eval:} Table \ref{tab:g_eval_scores} highlights GPT-4o's dominance in engagingness and naturalness with perfect scores (3.0) for DialogSum, while InternLM-2.5 stands out in coherence (2.9990) and groundedness (2.9973) for DialogSum, and coherence (2.9983) and groundedness (2.9952) for SAMSum, suggesting it maintains high factual accuracy.

\noindent \textbf{Dataset-Specific Performance.  }The contrasting performance of GPT-4o on the DialogSum and SAMSum datasets in Table \ref{tab:fed_scores} can be attributed to the differing structures of the dialogues in these datasets. DialogSum consists of more formal and structured dialogues, which aligns with the typical response style of GPT-4o, leading to its stronger performance. In contrast, SAMSum contains more casual, human-like conversations, which might explain GPT-4o's relatively poorer performance, as it may not adapt as well to the informal, spontaneous nature of such dialogues. Overall, while GPT-4o excels in natural and engaging dialogue, LLaMA-3 offers the most versatility, and InternLM-2.5 provides a strong alternative with high coherence and groundedness.

\begin{table}[htbp]
\centering
\resizebox{\columnwidth}{!}{%
    \begin{tabular}{lcccc}
    \toprule
     & \textbf{engagingness} & \textbf{naturalness} & \textbf{coherence} & \textbf{groundedness} \\
    \midrule
    \multicolumn{5}{c}{\textbf{DIALOGUESUM}} \\
    \midrule
    \textbf{Phi-3} & 2.5236 & 2.7238 & 2.6308 & 2.5557 \\
    \textbf{InternLM-2.5} & 2.9995 & 2.9989 & 2.9990 & 2.9973 \\
    \textbf{LLaMA-3} & 2.9987 & 2.9988 & 2.9972 & 2.9935 \\
    \textbf{GPT-4o} & \textbf{3} & \textbf{3} & \textbf{3} & \textbf{2.9975} \\
    \midrule
    \multicolumn{5}{c}{\textbf{SAMSUM}} \\
    \midrule
    \textbf{Phi-3} & 2.4623 & 2.6821 & 2.5848 & 2.5060 \\
    \textbf{InternLM-2.5} & 2.9992 & 2.9969 & \textbf{2.9983} & \textbf{2.9952} \\
    \textbf{LLaMA-3} & 2.9976 & 2.9971 & 2.9969 & 2.9916 \\
    \textbf{GPT-4o} & \textbf{2.9994} & \textbf{2.9977} & 2.9982 & 2.9944 \\
    \bottomrule
    \end{tabular}
}
\caption{G-EVAL}
\label{tab:g_eval_scores}
\end{table}

\subsubsection{Strong performance of LLaMA-3}

\begin{table*}[htbp]
\centering

\begin{tabular}{lccccccc}
\toprule
\textbf{Model} & \multicolumn{3}{c}{\textbf{Before Finetuning}} & \multicolumn{3}{c}{\textbf{Finetuning on In-Domain Data}} \\
\cmidrule(lr){2-4} \cmidrule(lr){5-7}
 & \textbf{QAGS} & \textbf{BERTScore} & \textbf{ROUGE-L} & \textbf{QAGS} & \textbf{BERTScore} & \textbf{ROUGE-L} \\
\midrule
\multicolumn{7}{c}{\textbf{DIALOGSUM}} \\
\midrule
\textbf{distillbart-cnn} & 0.6134 & 0.5093 & 0.1950 & 0.5586 & 0.7005 & 0.3367 \\
\textbf{bart-base} & 0.7007 & 0.5274 & 0.1375 & 0.4789 & 0.6868 & 0.2969 \\
\textbf{t5-base} & 0.5901 & 0.5491 & 0.1812 & 0.4766 & 0.6953 & 0.2986 \\
\textbf{led-base-16384} & 0.8261 & 0.5471 & 0.1634 & 0.4872 & 0.7084 & 0.3165 \\
\midrule
\multicolumn{7}{c}{\textbf{SAMSUM}} \\
\midrule
\textbf{distillbart-cnn} & 0.6627 & 0.5500 & 0.2394 & 0.6041 & 0.6849 & 0.3578 \\
\textbf{bart-base} & 0.7563 & 0.4389 & 0.1765 & 0.5302 & 0.6520 & 0.3049 \\
\textbf{t5-base} & 0.5574 & 0.4190 & 0.1237 & 0.5460 & 0.6448 & 0.3000 \\
\textbf{led-base-16384} & 0.7429 & 0.4310 & 0.1812 & 0.5440 & 0.6522 & 0.3175 \\
\bottomrule
\end{tabular}

\caption{Performance of models before and after finetuning on in-domain data}
\label{tab:finetuning_scores}
\end{table*}

The observed superiority of LLaMA-3 over GPT-4o is surprising because an 8 billion 8-bit quantized model not only competes with but also performs better than GPT-4o in certain metrics. We hypothesize that this could be due to the way GPT-4o was trained, which might make it more constrained in its responses, whereas LLaMA-3, being an open-source model, operates with fewer restrictions. This allows LLaMA-3 to exhibit greater flexibility, diversity, and adaptability in generating dialogues, potentially explaining its better performance in certain metrics. These characteristics can be seen in criteria like 'inquisitiveness' and 'likeability' in Table \ref{tab:gpt_scores} and, 'depth' and 'diverse' in Table \ref{tab:fed_scores}. These results suggest that for building human-like data generation frameworks, open-source LLMs are a more suitable choice than closed-source LLMs. The minimal constraints on response formatting during the training of open-source models enable them to generate more diverse, flexible, and human-like dialogues, making them better suited for tasks requiring natural and conversational interactions.

\subsection{Fine-tuning and Performance Results}

\begin{table*}[t]
\centering
\resizebox{\textwidth}{!}{
\begin{tabular}{l|ccc|ccc|ccc|ccc}
\toprule
\textbf{Model} & \multicolumn{3}{c}{\textbf{Phi-3}} & \multicolumn{3}{c}{\textbf{InternLM-2.5}} & \multicolumn{3}{c}{\textbf{LLaMA-3}} & \multicolumn{3}{c}{\textbf{GPT-4o}}\\
\cmidrule(lr){2-4} \cmidrule(lr){5-7} \cmidrule(lr){8-10} \cmidrule(lr){11-13}
 & \textbf{QAGS} & \textbf{BERTScore} & \textbf{ROUGE-L} & \textbf{QAGS} & \textbf{BERTScore} & \textbf{ROUGE-L} & \textbf{QAGS} & \textbf{BERTScore} & \textbf{ROUGE-L} & \textbf{QAGS} & \textbf{BERTScore} & \textbf{ROUGE-L}\\
\midrule
\multicolumn{13}{c}{\textbf{DIALOGUESUM}} \\
\midrule
\textbf{distillbart-cnn} & 0.6588 & 0.5778 & \textbf{0.2187} & 0.6420 & 0.6008 & 0.2167 & 0.6586 & 0.6161 & 0.2040 & \textbf{0.6713} & \textbf{0.6242} & 0.2014 \\

\textbf{bart-base} & 0.5355 & 0.5958 & \textbf{0.2029} & 0.5418 & \textbf{0.6212} & 0.1897 & \textbf{0.5825} & 0.6033 & 0.1789 & 0.5590 & 0.6039 & 0.1769\\

\textbf{t5-base} & 0.5937 & 0.5949 & \textbf{0.2047} & 0.5825 & 0.5941 & 0.1878 & 0.6034 & 0.6172 & 0.1959 & \textbf{0.6305} & \textbf{0.6319} & 0.2044\\

\textbf{led-base-16384} & 0.5358 & 0.6129 & \textbf{0.2109} & 0.5189 & 0.6027 & 0.1606 & 0.5697 & 0.6302 & 0.1999 & \textbf{0.5791} & \textbf{0.6308} & 0.1989\\
\midrule
\multicolumn{13}{c}{\textbf{SAMSUM}} \\
\midrule
\textbf{distillbart-cnn} & 0.6585 & 0.5931 & 0.2262 & 0.6388 & 0.6066 & 0.2422 & \textbf{0.6849} & \textbf{0.6029} & \textbf{0.2374} & 0.6757 & 0.6029 & 0.2291\\

\textbf{bart-base} & 0.5648 & 0.5665 & 0.2146 & 0.5435 & 0.5663 & 0.2021 & \textbf{0.6132} & \textbf{0.5899} & \textbf{0.2345} & 0.5707 & 0.5808 & 0.2154\\

\textbf{t5-base} & 0.5905 & 0.5397 & \textbf{0.2085} & 0.5457 & 0.5193 & 0.1854 & \textbf{0.6412} & 0.5054 & 0.1976 & 0.6023 & \textbf{0.5419} & 0.1979\\

\textbf{led-base-16384} & 0.5883 & 0.5477 & 0.2289 & 0.5457 & 0.5615 & 0.2167 & \textbf{0.5917} & \textbf{0.5785} & \textbf{0.2390} & 0.5738 & 0.569 & 0.2298\\
\bottomrule
\end{tabular}
}
\caption{Performance after finetuning on synthetic data}
\label{tab:model_performance_sdg}
\end{table*}

To validate the usability of the synthetic data generated using DiaSynth, we fine-tuned and evaluated several pretrained language models on the task of dialogue summarization. The summaries for dialogues generated by different LLMs were created using the corresponding LLMs through prompting. The pretrained models used for evaluation include DistilBART, BART, T5, and LED.

The experimental setup is designed as follows:

\begin{itemize} 
    \item Metrics are reported on the validation and test sets of DialogSum and SAMSum. 
    \item To evaluate DiaSynth-generated data, we compared models fine-tuned on DiaSynth data with their base versions (no fine-tuning). 
    \item In-domain training sets were randomly sampled to match the size of the DiaSynth-generated data, enabling fair comparison. 
    \item The experiment aimed to quantify performance improvement of DiaSynth-fine-tuned models and assess alignment with in-domain data distributions. \item Models were fine-tuned for 2 epochs with a learning rate of \texttt{5e-5} and a warmup of \texttt{50} steps. 
\end{itemize}

The results presented in Tables \ref{tab:finetuning_scores} and \ref{tab:model_performance_sdg} present the performance of the base models, models finetuned on in-domain data and models finetuned on DiaSynth generated data. Models finetuned on DiaSynth data generally improves the performances from the BERTScore and ROUGE-L metrics. Surprisingly, for some models (LED and BART) the QAGS scores were higher than the models finetuned on DiaSynth. On further exploration, we found out that these models extracted multiple sentences from the given dialogue instead of generating a summary which led to high QAGS scores. Comparing models finetuned on in-domain data to those finetuned on DiaSynth data reveals that DiaSynth finetuning generally enhances factual accuracy, with BERTScore and ROUGE-L scores remaining comparable. The disparity in BERTScore and ROUGE-L results may be due to format variations. Models fine-tuned on in-domain data were evaluated on summaries that matched the training format closely, while DiaSynth-fine-tuned models were trained on LLM-generated summaries and evaluated on human-generated summaries, leading to minor format mismatches. Comparison between the different LLMs from Table \ref{tab:model_performance_sdg}, shows that GPT-4o is better at generating dialogues and summaries that are formal in nature while LLaMA-3 and open source LLMs would be better for generating dialogues that are informal and casual in nature.

\begin{subequations}
\begin{flalign}
\text{\% Improvement} & = \frac{\text{After - Before}}{\text{Before}} \label{eq:percentage_improvement}\\
\text{\% Coverage} & = \frac{\text{Score DiaSynth}}{\text{Score In-domain}} \label{eq:coverage_percentage}
\end{flalign}
\end{subequations}

\begin{table}[htbp]
    \centering
    \begin{tabular}{ccc}
         \hline
         \textbf{Model} & \textbf{\% Improvement} & \textbf{\% Covered} \\
         \hline
         distilbart-cnn & 10.96 & 88.81  \\
         bart-base & 9.21 & 90.6 \\
         t5-base & 7.59 &  93.67 \\
         led-base-16384 & 2.14 & 89.68 \\
         \hline
    \end{tabular}
    \caption{Summarization results on DialogSum}
    \label{tab:dialogsum_improvement_covered}
\end{table}

\begin{table}[htbp]
    \centering
    \begin{tabular}{ccc}
         \hline
         \textbf{Model} & \textbf{\% Improvement} & \textbf{\% Covered} \\
         \hline
         distilbart-cnn & 6.07 & 87.25  \\
         bart-base & 16.12 & 94.35 \\
         t5-base & 30.04 &  87.36 \\
         led-base-16384 & 15.25 & 90.91 \\
         \hline
    \end{tabular}
    \caption{Summarization results on SAMSum}
    \label{tab:samsum_improvement_covered}
\end{table}

To assess the percentage improvement and  percentage coverage of the distributional characteristics of the in-domain data by the synthetically generated data, we use Equations \ref{eq:percentage_improvement} and \ref{eq:coverage_percentage} respectively. We use the scores of models finetuned on LLaMA-3 generated data because of its dominance in both quality and usability. Across the 24 reported results, the overall coverage percentage of the LLaMA-3 generated data is \textbf{90.48\%}. Notably, the QAGS scores of models fine-tuned on synthetic data surpass those of models trained on in-domain data, suggesting that synthetic data can match or even exceed in-domain data performance in some aspects. Excluding QAGS, the coverage percentage is calculated to be \textbf{77.07\%}. In addition to the average percentages, we also present the model wise percentage improvement and coverage in Table \ref{tab:dialogsum_improvement_covered} and \ref{tab:samsum_improvement_covered}. The results presented are with respect to the dialogues generated using LLaMA-3 and they illustrate clear improvements for every model, highlighting that even with moderate LLMs of small scale (3B - 8B), high-quality synthetic dialogue datasets can be effectively created across different domains and different dialogue formats. 

\section{Conclusion}
\label{sec:conclusion}
In this paper, we introduced DiaSynth, a synthetic dialogue generation framework capable of producing high-quality, contextually rich dialogues across a wide range of domains. Our experiments demonstrated that models fine-tuned on DiaSynth-generated data exhibit significant improvements in downstream tasks, as evidenced by substantial increases in BERTScore and ROUGE-L compared to their base models. These results highlight the potential of DiaSynth as an effective tool for generating dialogue data, particularly in domains where training data is scarce.

Furthermore, our analysis showed that different LLMs excel in different dialogue structures, with LLaMA-3 performing better for informal dialogues and GPT-4o for more structured settings. This insight suggests that leveraging open-source LLMs may be more advantageous for generating human-like conversational data. Despite certain limitations, such as varying LLM performance across dialogue types and knowledge gaps in zero-shot generation, DiaSynth presents a promising approach to dialogue data generation and offers a valuable resource for future advancements in building more sophisticated and adaptable dialogue systems.

\section{Limitations}
\label{sec:limitations}
Despite the promising results, our approach has some limitations. Firstly, different LLMs exhibit varied performance based on the dialogue structure, with certain models like LLaMA-3 performing better for more informal dialogues (e.g., SAMSum) and others like GPT-4o excelling in structured, formal dialogues (e.g., DialogSum). This indicates that there is no single model that can universally handle all types of dialogue structures, but a single SOTA model can give stable and decent results like LLaMA-3.

Secondly, the generation process may suffer from a lack of knowledge on certain topics, especially in cases where the LLMs were not sufficiently trained on those domains. Additionally, our framework relies on zero-shot generation for personas and sub-topics, which, while flexible, can sometimes result in less coherent or less accurate persona simulation, as it is not fine-tuned for specific contexts. 

Since LLMs power DiaSynth, hallucinations and compute-need are two inherent limitations. We present a detailed hallucination study in Appendix \ref{app:hallucination}, which indicates that though the hallucinations rates are acceptable, there is still scope for improvements. These limitations suggest directions for future work, such as combining LLMs to leverage their strengths or incorporating more topic-specific training to enhance knowledge coverage.

\bibliography{custom}

\begin{thebibliography}{24}
\providecommand{\natexlab}[1]{#1}

\bibitem[{Abdin et~al.(2024)Abdin, Jacobs, Awan, Aneja, Awadallah, Awadalla, Bach, Bahree, Bakhtiari, Behl et~al.}]{abdin2024phi3}
Marah Abdin, Sam~Ade Jacobs, Ammar~Ahmad Awan, Jyoti Aneja, Ahmed Awadallah, Hany Awadalla, Nguyen Bach, Amit Bahree, Arash Bakhtiari, Harkirat Behl, et~al. 2024.
\newblock Phi-3 technical report: A highly capable language model locally on your phone.
\newblock \emph{arXiv preprint arXiv:2404.14219}.

\bibitem[{Beltagy et~al.(2020)Beltagy, Peters, and Cohan}]{beltagy2020longformer}
Iz~Beltagy, Matthew~E Peters, and Arman Cohan. 2020.
\newblock Longformer: The long-document transformer.
\newblock \emph{arXiv preprint arXiv:2004.05150}.

\bibitem[{Budzianowski et~al.(2018)Budzianowski, Wen, Tseng, Casanueva, Ultes, Ramadan, and Ga{\v{s}}i{\'c}}]{budzianowski-etal-2018-multiwoz}
Pawe{\l} Budzianowski, Tsung-Hsien Wen, Bo-Hsiang Tseng, I{\~n}igo Casanueva, Stefan Ultes, Osman Ramadan, and Milica Ga{\v{s}}i{\'c}. 2018.
\newblock \href {https://doi.org/10.18653/v1/D18-1547} {{M}ulti{WOZ} - a large-scale multi-domain {W}izard-of-{O}z dataset for task-oriented dialogue modelling}.
\newblock In \emph{Proceedings of the 2018 Conference on Empirical Methods in Natural Language Processing}, pages 5016--5026, Brussels, Belgium. Association for Computational Linguistics.

\bibitem[{Cai et~al.(2024)Cai, Cao, Chen, Chen, Chen, Chen, Chen, Chen, Chen, Chu, wen Dong, Duan, Fan, Fei, Gao, Ge, Gu, Gu, Gui, Guo, Guo, He, Hu, Huang, Jiang, Jiao, Jin, Lei, Li, Li, Li, Li, Li, Li, Liu, Liu, Hong, Liu, Liu, Liu, Lv, Lv, Lv, Ma, Ma, Ma, Ning, Ouyang, Qiu, Qu, Shang, Shao, Song, Song, Sui, Sun, Sun, Tang, Wang, Wang, Wang, Wang, Wang, Wang, Wang, Wei, Weng, Wu, Xiong, Xu, Xu, Yan, Yan, Yang, Ye, Ying, Yu, Yu, Zang, Zhang, Zhang, Zhang, Zhang, Zhang, Zhang, Zhang, Zhang, Zhang, Zhang, Zhang, Zhao, Zhao, Zhao, Zhou, Zhou, Zhuo, Zou, Qiu, Qiao, and Lin}]{Cai2024InternLM2TR}
Zheng Cai, Maosong Cao, Haojiong Chen, Kai Chen, Keyu Chen, Xin Chen, Xun Chen, Zehui Chen, Zhi Chen, Pei Chu, Xiao wen Dong, Haodong Duan, Qi~Fan, Zhaoye Fei, Yang Gao, Jiaye Ge, Chenya Gu, Yuzhe Gu, Tao Gui, Aijia Guo, Qipeng Guo, Conghui He, Yingfan Hu, Ting Huang, Tao Jiang, Penglong Jiao, Zhen Jin, Zhikai Lei, Jiaxing Li, Jingwen Li, Linyang Li, Shuaibin Li, Wei Li, Yining Li, Hongwei Liu, Jiangning Liu, Jiawei Hong, Kaiwen Liu, Kui-Jie Liu, Xiaoran Liu, Chen Lv, Haijun Lv, Kai Lv, Li~Ma, Runyuan Ma, Zerun Ma, Wenchang Ning, Linke Ouyang, Jiantao Qiu, Yuan Qu, Fukai Shang, Yunfan Shao, Demin Song, Zifan Song, Zhihao Sui, Peng Sun, Yu~Sun, Huanze Tang, Bin Wang, Guoteng Wang, Jiaqi Wang, Jiayu Wang, Rui Wang, Yudong Wang, Ziyi Wang, Xing Wei, Qizhen Weng, Fan Wu, Yingtong Xiong, Chao Xu, Rui~Ze Xu, Hang Yan, Yirong Yan, Xiaogui Yang, Haochen Ye, Huaiyuan Ying, Jia Yu, Jing Yu, Yuhang Zang, Chuyu Zhang, Li~Zhang, Pan Zhang, Peng Zhang, Ruijie Zhang, Shuo Zhang, Songyang Zhang, Wenjian Zhang, Wenwei Zhang, Xingcheng Zhang, Xinyue Zhang, Hui Zhao, Qian Zhao, Xiaomeng Zhao, Fen-Fang Zhou, Zaida Zhou, Jingming Zhuo, Yi-Ling Zou, Xipeng Qiu, Yu~Qiao, and Dahua Lin. 2024.
\newblock \href {https://api.semanticscholar.org/CorpusID:268691939} {Internlm2 technical report}.
\newblock \emph{ArXiv}, abs/2403.17297.

\bibitem[{Chan et~al.(2024)Chan, Wang, Yu, Mi, and Yu}]{chan2024scaling}
Xin Chan, Xiaoyang Wang, Dian Yu, Haitao Mi, and Dong Yu. 2024.
\newblock Scaling synthetic data creation with 1,000,000,000 personas.
\newblock \emph{arXiv preprint arXiv:2406.20094}.

\bibitem[{Chen et~al.(2021)Chen, Liu, Chen, and Zhang}]{chen-etal-2021-dialogsum}
Yulong Chen, Yang Liu, Liang Chen, and Yue Zhang. 2021.
\newblock \href {https://doi.org/10.18653/v1/2021.findings-acl.449} {{D}ialog{S}um: {A} real-life scenario dialogue summarization dataset}.
\newblock In \emph{Findings of the Association for Computational Linguistics: ACL-IJCNLP 2021}, pages 5062--5074, Online. Association for Computational Linguistics.

\bibitem[{Dubey et~al.(2024)Dubey, Jauhri, Pandey, Kadian, Al-Dahle, Letman, Mathur, Schelten, Yang, Fan et~al.}]{dubey2024llama}
Abhimanyu Dubey, Abhinav Jauhri, Abhinav Pandey, Abhishek Kadian, Ahmad Al-Dahle, Aiesha Letman, Akhil Mathur, Alan Schelten, Amy Yang, Angela Fan, et~al. 2024.
\newblock The llama 3 herd of models.
\newblock \emph{arXiv preprint arXiv:2407.21783}.

\bibitem[{Feng et~al.(2020)Feng, Wan, Gunasekara, Patel, Joshi, and Lastras}]{feng-etal-2020-doc2dial}
Song Feng, Hui Wan, Chulaka Gunasekara, Siva Patel, Sachindra Joshi, and Luis Lastras. 2020.
\newblock \href {https://doi.org/10.18653/v1/2020.emnlp-main.652} {doc2dial: A goal-oriented document-grounded dialogue dataset}.
\newblock In \emph{Proceedings of the 2020 Conference on Empirical Methods in Natural Language Processing (EMNLP)}, pages 8118--8128, Online. Association for Computational Linguistics.

\bibitem[{Friel and Sanyal(2023)}]{Friel2023ChainpollAH}
Robert Friel and Atindriyo Sanyal. 2023.
\newblock \href {https://api.semanticscholar.org/CorpusID:264590664} {Chainpoll: A high efficacy method for llm hallucination detection}.
\newblock \emph{ArXiv}, abs/2310.18344.

\bibitem[{Fu et~al.(2024)Fu, Ng, Jiang, and Liu}]{fu-etal-2024-gptscore}
Jinlan Fu, See-Kiong Ng, Zhengbao Jiang, and Pengfei Liu. 2024.
\newblock \href {https://doi.org/10.18653/v1/2024.naacl-long.365} {{GPTS}core: Evaluate as you desire}.
\newblock In \emph{Proceedings of the 2024 Conference of the North American Chapter of the Association for Computational Linguistics: Human Language Technologies (Volume 1: Long Papers)}, pages 6556--6576, Mexico City, Mexico. Association for Computational Linguistics.

\bibitem[{Gliwa et~al.(2019)Gliwa, Mochol, Biesek, and Wawer}]{gliwa-etal-2019-samsum}
Bogdan Gliwa, Iwona Mochol, Maciej Biesek, and Aleksander Wawer. 2019.
\newblock \href {https://doi.org/10.18653/v1/D19-5409} {{SAMS}um corpus: A human-annotated dialogue dataset for abstractive summarization}.
\newblock In \emph{Proceedings of the 2nd Workshop on New Frontiers in Summarization}, pages 70--79, Hong Kong, China. Association for Computational Linguistics.

\bibitem[{Han et~al.(2024)Han, Koh, Seo, Chang, and Sohn}]{han2024psydial}
Ji-Eun Han, Jun-Seok Koh, Hyeon-Tae Seo, Du-Seong Chang, and Kyung-Ah Sohn. 2024.
\newblock \href {https://aclanthology.org/2024.lrec-main.1166} {{PSYDIAL}: Personality-based synthetic dialogue generation using large language models}.
\newblock In \emph{Proceedings of the 2024 Joint International Conference on Computational Linguistics, Language Resources and Evaluation (LREC-COLING 2024)}, pages 13321--13331, Torino, Italia. ELRA and ICCL.

\bibitem[{Lewis et~al.(2020)Lewis, Liu, Goyal, Ghazvininejad, Mohamed, Levy, Stoyanov, and Zettlemoyer}]{lewis-etal-2020-bart}
Mike Lewis, Yinhan Liu, Naman Goyal, Marjan Ghazvininejad, Abdelrahman Mohamed, Omer Levy, Veselin Stoyanov, and Luke Zettlemoyer. 2020.
\newblock \href {https://doi.org/10.18653/v1/2020.acl-main.703} {{BART}: Denoising sequence-to-sequence pre-training for natural language generation, translation, and comprehension}.
\newblock In \emph{Proceedings of the 58th Annual Meeting of the Association for Computational Linguistics}, pages 7871--7880, Online. Association for Computational Linguistics.

\bibitem[{Lin(2004)}]{lin-2004-rouge}
Chin-Yew Lin. 2004.
\newblock \href {https://aclanthology.org/W04-1013} {{ROUGE}: A package for automatic evaluation of summaries}.
\newblock In \emph{Text Summarization Branches Out}, pages 74--81, Barcelona, Spain. Association for Computational Linguistics.

\bibitem[{Liu et~al.(2023)Liu, Iter, Xu, Wang, Xu, and Zhu}]{liu-etal-2023-g}
Yang Liu, Dan Iter, Yichong Xu, Shuohang Wang, Ruochen Xu, and Chenguang Zhu. 2023.
\newblock \href {https://doi.org/10.18653/v1/2023.emnlp-main.153} {{G}-eval: {NLG} evaluation using gpt-4 with better human alignment}.
\newblock In \emph{Proceedings of the 2023 Conference on Empirical Methods in Natural Language Processing}, pages 2511--2522, Singapore. Association for Computational Linguistics.

\bibitem[{Manakul et~al.(2023)Manakul, Liusie, and Gales}]{manakul-etal-2023-selfcheckgpt}
Potsawee Manakul, Adian Liusie, and Mark Gales. 2023.
\newblock \href {https://doi.org/10.18653/v1/2023.emnlp-main.557} {{S}elf{C}heck{GPT}: Zero-resource black-box hallucination detection for generative large language models}.
\newblock In \emph{Proceedings of the 2023 Conference on Empirical Methods in Natural Language Processing}, pages 9004--9017, Singapore. Association for Computational Linguistics.

\bibitem[{Mehri and Eskenazi(2020)}]{mehri-eskenazi-2020-unsupervised}
Shikib Mehri and Maxine Eskenazi. 2020.
\newblock \href {https://doi.org/10.18653/v1/2020.sigdial-1.28} {Unsupervised evaluation of interactive dialog with {D}ialo{GPT}}.
\newblock In \emph{Proceedings of the 21th Annual Meeting of the Special Interest Group on Discourse and Dialogue}, pages 225--235, 1st virtual meeting. Association for Computational Linguistics.

\bibitem[{Raffel et~al.(2020)Raffel, Shazeer, Roberts, Lee, Narang, Matena, Zhou, Li, and Liu}]{raffel2020exploring}
Colin Raffel, Noam Shazeer, Adam Roberts, Katherine Lee, Sharan Narang, Michael Matena, Yanqi Zhou, Wei Li, and Peter~J Liu. 2020.
\newblock Exploring the limits of transfer learning with a unified text-to-text transformer.
\newblock \emph{Journal of machine learning research}, 21(140):1--67.

\bibitem[{Steindl et~al.(2023)Steindl, Sch\"{a}fer, and Ludwig}]{cheng2024generating}
Sebastian Steindl, Ulrich Sch\"{a}fer, and Bernd Ludwig. 2023.
\newblock \href {https://doi.org/10.1007/978-3-031-42608-7_17} {Generating synthetic dialogues from prompts to improve task-oriented dialogue systems}.
\newblock In \emph{KI 2023: Advances in Artificial Intelligence: 46th German Conference on AI, Berlin, Germany, September 26–29, 2023, Proceedings}, page 207–214, Berlin, Heidelberg. Springer-Verlag.

\bibitem[{Wang et~al.(2020)Wang, Cho, and Lewis}]{qags}
Alex Wang, Kyunghyun Cho, and Mike Lewis. 2020.
\newblock \href {https://doi.org/10.18653/v1/2020.acl-main.450} {Asking and answering questions to evaluate the factual consistency of summaries}.
\newblock In \emph{Proceedings of the 58th Annual Meeting of the Association for Computational Linguistics}, pages 5008--5020, Online. Association for Computational Linguistics.

\bibitem[{Wang et~al.(2022)Wang, Kordi, Mishra, Liu, Smith, Khashabi, and Hajishirzi}]{wang2022self}
Yizhong Wang, Yeganeh Kordi, Swaroop Mishra, Alisa Liu, Noah~A Smith, Daniel Khashabi, and Hannaneh Hajishirzi. 2022.
\newblock Self-instruct: Aligning language models with self-generated instructions.
\newblock \emph{arXiv preprint arXiv:2212.10560}.

\bibitem[{Wei et~al.(2024)Wei, Wang, Schuurmans, Bosma, Ichter, Xia, Chi, Le, and Zhou}]{cot}
Jason Wei, Xuezhi Wang, Dale Schuurmans, Maarten Bosma, Brian Ichter, Fei Xia, Ed~H. Chi, Quoc~V. Le, and Denny Zhou. 2024.
\newblock Chain-of-thought prompting elicits reasoning in large language models.
\newblock In \emph{Proceedings of the 36th International Conference on Neural Information Processing Systems}, NIPS '22, Red Hook, NY, USA. Curran Associates Inc.

\bibitem[{Zeng et~al.(2020)Zeng, Yang, Ju, Yang, Wang, Zhang, Zhou, Zeng, Dong, Zhang, Fang, Zhu, Chen, and Xie}]{zeng-etal-2020-meddialog}
Guangtao Zeng, Wenmian Yang, Zeqian Ju, Yue Yang, Sicheng Wang, Ruisi Zhang, Meng Zhou, Jiaqi Zeng, Xiangyu Dong, Ruoyu Zhang, Hongchao Fang, Penghui Zhu, Shu Chen, and Pengtao Xie. 2020.
\newblock \href {https://doi.org/10.18653/v1/2020.emnlp-main.743} {{M}ed{D}ialog: Large-scale medical dialogue datasets}.
\newblock In \emph{Proceedings of the 2020 Conference on Empirical Methods in Natural Language Processing (EMNLP)}, pages 9241--9250, Online. Association for Computational Linguistics.

\bibitem[{Zhang* et~al.(2020)Zhang*, Kishore*, Wu*, Weinberger, and Artzi}]{Zhang*2020BERTScore:}
Tianyi Zhang*, Varsha Kishore*, Felix Wu*, Kilian~Q. Weinberger, and Yoav Artzi. 2020.
\newblock \href {https://openreview.net/forum?id=SkeHuCVFDr} {Bertscore: Evaluating text generation with bert}.
\newblock In \emph{International Conference on Learning Representations}.

\end{thebibliography}
\appendix

\section{Hallucination Study}
\label{app:hallucination}

In addition to evaluating the quality and usability of dialogues produced by DiaSynth, 
we conducted a study on the phenomenon of hallucinations within the generated dialogues. 
Hallucinations in language models refer to instances where the output contains misleading or 
incorrect information or situations where the model repeats the same content. To evaluate the 
occurrence of hallucinations, we compared the generated dialogues with their respective summaries 
and assessed them using two well-known hallucination benchmarks: \textbf{SelfCheckGPT} 
\cite{manakul-etal-2023-selfcheckgpt} and \textbf{ChainPoll} \cite{Friel2023ChainpollAH}. 
This analysis provides insights into the prevalence of hallucinations and informs strategies 
for improving dialogue quality in future iterations of DiaSynth. The results are presented in 
Tables \ref{tab:hallu_ds} and \ref{tab:hallu_ss}. 
 
\subsection{SelfCheckGPT}
SelfCheckGPT quantifies the self-consistency of LLM outputs by examining agreement across multiple 
outputs from the same prompt. This assessment reveals potential inaccuracies through metrics like 
SelfCheck-BertScore.

The SelfCheck-BERTScore results for various models show that hallucination levels are at worst around 
25\%, which is acceptable but still indicates areas for improvement. Across both datasets, 
\textbf{Phi-3} demonstrates the most robustness, likely due to its pretraining on structured, 
textbook-like data, which may contribute to greater consistency and factual accuracy.

 \subsection{ChainPoll}
ChainPoll utilizes a chain-of-thought prompting approach to identify hallucinations by iteratively 
polling the model with structured reasoning prompts. This method systematically detects both open-domain 
and closed-domain hallucinations, where lower scores indicate fewer hallucinations.

The ChainPoll scores indicate that hallucination levels on these models are generally low, with 
the best performance seen by \textbf{GPT-4o} on SAMSum, which achieves the lowest score of 0.120, 
suggesting minimal hallucinations. On the other hand, \textbf{LLaMA-3} scores higher at 0.237 on 
SAMSum, indicating more frequent hallucinations. These findings highlight different models' strengths 
in generating accurate and reliable dialogues.

\subsection{Implications for DiaSynth}
The results from both SelfCheckGPT and ChainPoll evaluations suggest that DiaSynth, when leveraging 
models like Llama 3, is capable of generating dialogues with relatively low hallucination rates. 
However, specific models show variability in performance across datasets, indicating that further 
enhancements, such as fine-tuning or incorporating additional guardrails, could improve DiaSynth’s 
robustness in generating reliable dialogues across diverse domains.

\begin{table}[htbp]
    \centering
    \resizebox{\columnwidth}{!}{%
        \begin{tabular}{lcc}
        \hline
        \textbf{LLM} & \textbf{ChainPoll} & \textbf{SCGPT-BERTScore} \\
        \hline
        Phi-3 & 0.198 & 0.791 \\
        InternLM-2.5 & 0.199 & 0.726 \\
        LLaMA-3 & 0.205 & 0.793 \\
        GPT-4o & 0.178 & 0.765 \\
        \hline
        \end{tabular}
    }
    \caption{Hallucination calculation for DialogSum few-shot data}
    \label{tab:hallu_ds}
\end{table}

\begin{table}[htbp]
    \centering
    \resizebox{\columnwidth}{!}{%
        \begin{tabular}{lcc}
        \hline
        \textbf{LLM} & \textbf{ChainPoll} & \textbf{SCGPT-BERTScore} \\
        \hline
        Phi-3 & 0.154 & 0.785 \\
        InternLM-2.5 & 0.159 & 0.716 \\
        LLaMA-3 & 0.237 & 0.733 \\
        GPT-4o & 0.120 & 0.742 \\
        \hline
        \end{tabular}
    }
    \caption{Hallucination calculation for SAMSum few-shot data}
    \label{tab:hallu_ss}
\end{table}

\clearpage

\section{Scalability of DiaSynth}
\label{app:scalability}
This section illustrates the scalability of DiaSynth with a python program and examples.

\lstset{
  language=Python,
  basicstyle=\ttfamily\footnotesize,
  keywordstyle=\color{blue}\bfseries,
  commentstyle=\color{green!50!black},
  stringstyle=\color{red!70!black},
  showstringspaces=false,
  numbers=left,
  numberstyle=\tiny\color{gray},
  stepnumber=1,
  numbersep=10pt,  % Increase this from 5pt to 10pt to push numbers away
  backgroundcolor=\color{gray!10},
  frame=single,
  breaklines=true,
  breakatwhitespace=true,
  tabsize=4,
  captionpos=b
}

\begin{algorithm}
    \caption{Calculation of Total Dialogs Generated}
    \label{alg:total_dialogs}
    \begin{algorithmic}[1]
    \Procedure{CalculateTotalDialogs}{n, m, p}
        \State $ \text{total\_subtopics} \gets n \times m $
        \State $ \text{dialogs\_per\_subtopic} \gets \frac{p \times (p - 1)}{2} $
        \State $ \text{total\_dialogs} \gets \text{total\_subtopics} \times \text{dialogs\_per\_subtopic} $
        \State \textbf{return} total\_dialogs
    \EndProcedure
    \end{algorithmic}
    \end{algorithm}

\subsection{Example 1}
\begin{itemize}
    \item Number of topics (\( n \)): 10
    \item Number of subtopics per topic (\( m \)): 5
    \item Number of personas per subtopic (\( p \)): 3
\end{itemize}

\textbf{Calculation:}
\begin{align*}
\text{Total subtopics} &= n \times m = 10 \times 5 = 50 \\
\text{Dialogues per subtopic} &= \frac{p \times (p - 1)}{2} = \frac{3 \times 2}{2} = 3 \\
\text{Total dialogues} &= 50 \times 3 = 150
\end{align*}

This setup generates \textbf{150 dialogues}.

\subsection{Example 2}
\begin{itemize}
    \item Number of topics (\( n \)): 20
    \item Number of subtopics per topic (\( m \)): 4
    \item Number of personas per subtopic (\( p \)): 5
\end{itemize}

\textbf{Calculation:}
\begin{align*}
\text{Total subtopics} &= 20 \times 4 = 80 \\
\text{Dialogues per subtopic} &= \frac{5 \times 4}{2} = 10 \\
\text{Total dialogues} &= 80 \times 10 = 800
\end{align*}

This setup generates \textbf{800 dialogues}.

\subsection{Example 3}
\begin{itemize}
    \item Number of topics (\( n \)): 15
    \item Number of subtopics per topic (\( m \)): 6
    \item Number of personas per subtopic (\( p \)): 10
\end{itemize}

\textbf{Calculation:}
\begin{align*}
\text{Total subtopics} &= 15 \times 6 = 90 \\
\text{Dialogues per subtopic} &= \frac{10 \times 9}{2} = 45 \\
\text{Total dialogues} &= 90 \times 45 = 4050
\end{align*}

This setup generates \textbf{4050 dialogues}.

\subsection{Scaling Observations}
\begin{itemize}
    \item \textbf{Linear Scaling with Topics and Subtopics}: Increasing the number of topics or subtopics results in a linear 
    increase in the total number of dialogues, making it straightforward to expand the scope of dialogue generation.
    \item \textbf{Exponential Scaling with Personas}: The number of dialogues scales exponentially as the number of personas 
    increases because each additional persona allows for more combinations, making the framework highly scalable for complex scenarios.
    \item \textbf{Practical Use Case}: For specific domains like academic, healthcare, or business, these parameters can be adjusted 
    to generate thousands of dialogues to fit the needs of various applications such as training chatbots, virtual assistants, or 
    dialogue-based assessments.
\end{itemize}

These examples illustrate DiaSynth's potential for rapid and scalable generation of dialogues, 
which can be tailored to different domains by simply adjusting the input parameters.

\clearpage

\section{Characteristics for the conversation}
\label{app:conv_characteristics}

Table \ref{tab:dialogue_characteristics} shows different characteristics that we let the LLMs 
reason and decide using CoT. Before generating the dialogues, the LLMs are prompted to first reason 
about the various characteristics list for the dialogue given the topic and the personas and how 
the LLMs reason are illustrated in Appendix \ref{app:cot_examples}. 

\begin{table}[htbp]
\centering
\begin{tabular}{p{0.35\columnwidth}p{0.55\columnwidth}}
\hline
\textbf{Characteristic} & \textbf{Description} \\
\hline
Age and Gender & Defines demographic details, influencing style and tone. \\

Familiarity Level & Affects formality and depth based on relationship between speakers. \\

Emotional States & Impacts tone and flow based on emotions (e.g., happy, sad). \\

Formality Level & Determines level of politeness or casualness. \\

Duration of the Conversation & Suggests the intended length and complexity of dialogue. \\

Communication Medium & Defines the medium (e.g., face-to-face, phone), influencing style. \\

Topic of the Conversation & Guides the content and direction of the dialogue. \\

Location of the Conversation & Adds context influencing formality and content. \\

Agreement or Disagreement & Drives dialogue dynamics based on agreement level. \\

Natural Dialogue Features & Adds authenticity with fillers, pauses, and slang. \\
\hline
\end{tabular}
\caption{Characteristics of the Dialogue for CoT Prompt}
\label{tab:dialogue_characteristics}
\end{table}

\clearpage

\section{Example CoT environments}
\label{app:cot_examples}

\begin{figure}[htbp]
    % \centering
    \includegraphics[width=\columnwidth]{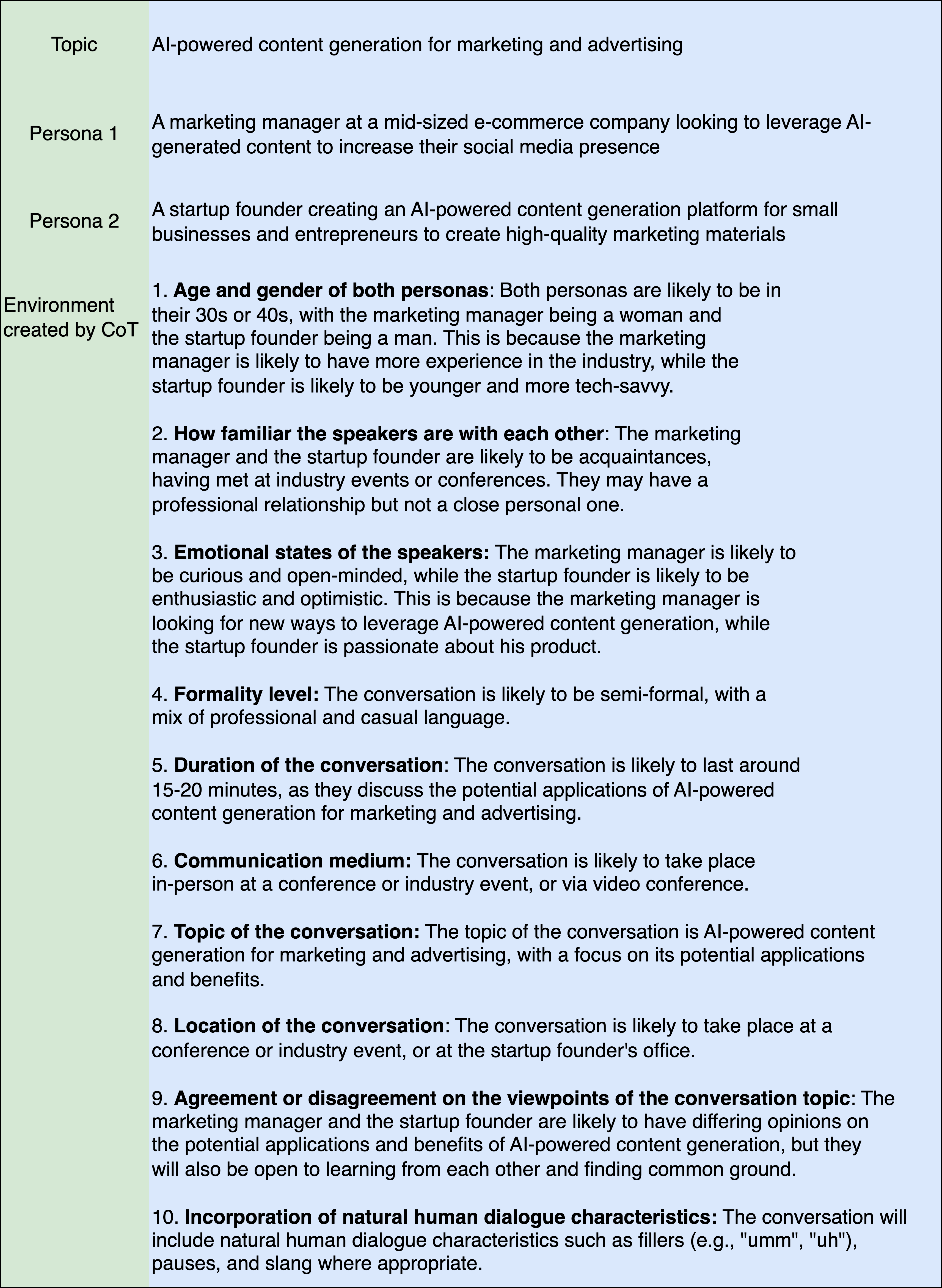}
    \caption{CoT Example 1}
    \label{fig:cot_example_1}
\end{figure}

\begin{figure}[H]
    % \centering
    \includegraphics[width=\columnwidth]{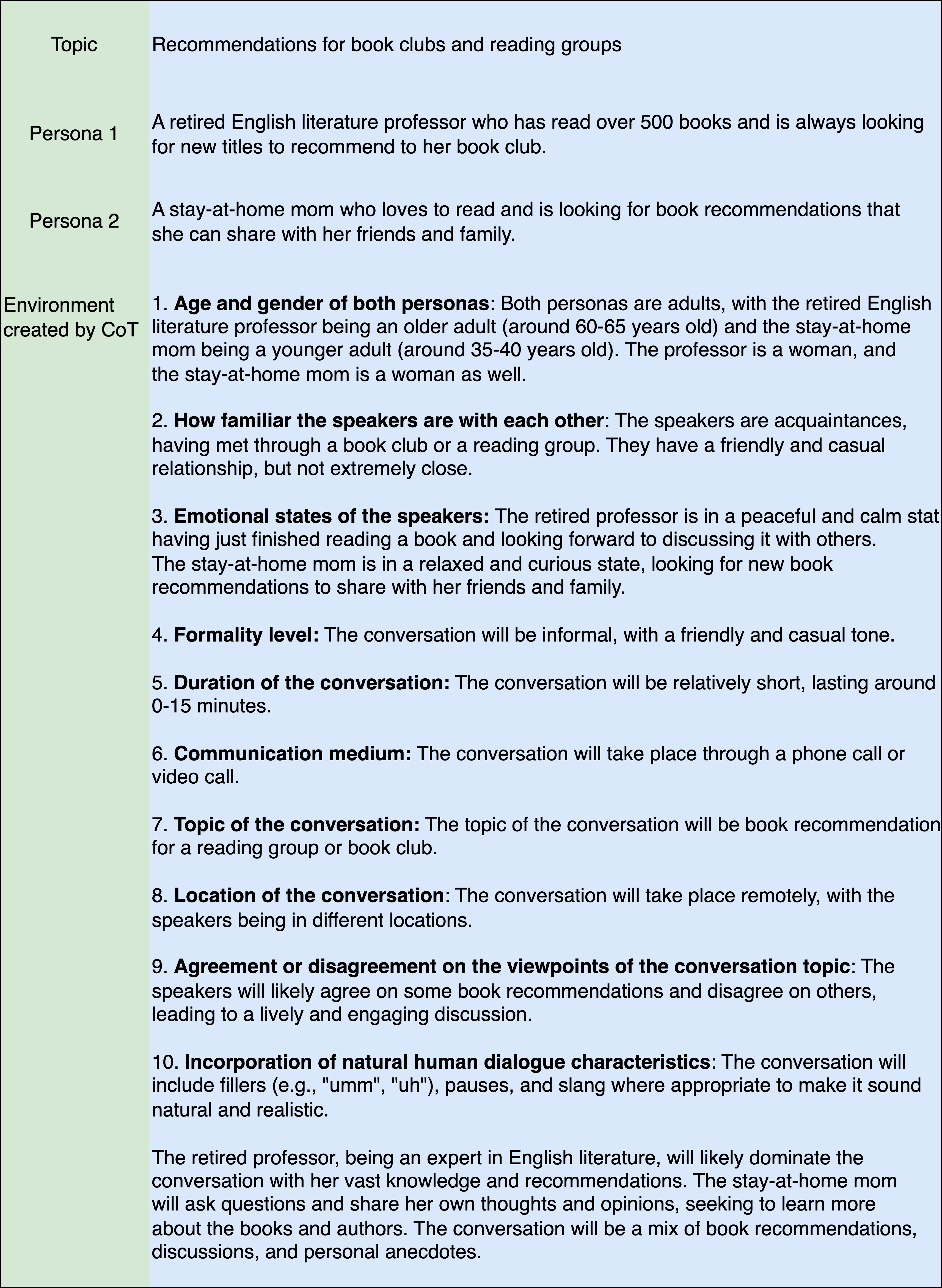}
    \caption{CoT Example 2}
    \label{fig:cot_example_2}
\end{figure}

These examples illustrate how CoT sets the various dialogue characteristics defined in 
Appendix \ref{app:conv_characteristics}. As can be seen in the examples, the characteristics for 
the conversation are completely changed based on the personas and the topics, resulting in more 
grounded conversation generation. Future works can further explore how CoT can be used to further 
break down to generate even more realistic dialogues. 
\newpage

\section{Ablation study on the use of CoT}
\label{app:ablation_cot}
In this section of the appendix, we present the impact of CoT in the dialogues generated by DiaSynth. 
We generate two datasets without CoT, using Phi-3 using DialogSum and SAMSum as few-shot examples with 
8 topics. Tables \ref{tab:ablation_cot_dialogsum_fed}, \ref{tab:ablation_cot_dialogsum_gptscore}, 
\ref{tab:ablation_cot_samsum_fed} and \ref{tab:ablation_cot_samsum_gptscore} compare the scores of 
dialogues generated with and without CoT for the FED score and GPTScore and it can be clearly that 
CoT tends to increase the quality of the dialogues generated by DiaSynth. 

We hypothesize that this improvement in quality is due to allowing the LLM to set diverse characteristics 
for the dialogue before generating the dialogue. This illustrates that either manually setting the 
relevant context or letting the LLM on its own to set the relevant context, we get better outputs, 
as adding relevant context lowers the probabilities of sequences that are not useful. The lower 
FED scores of CoT generated dialogues in Table \ref{tab:ablation_cot_samsum_fed}, might be because 
of the CoT generated dialogues being longer in length but it needs further research.

\begin{table}[H]
\centering
\begin{tabular}{lcc}
\hline
\textbf{Criteria} & \textbf{Without CoT} & \textbf{With CoT} \\ \hline
Coherent & 0.9507 & \textbf{0.9521} \\ 
Error Recovery & 0.938 & \textbf{0.9424}\\ 
Consistent & 0.9424 & \textbf{0.9523}\\ 
Diverse & 0.9431 & \textbf{0.952}\\ 
Depth & 0.9451 & \textbf{0.9506}\\ 
Likeable & \textbf{0.0088} & -0.0003\\ 
Understand & 0.9317 & \textbf{0.9338}\\ 
Flexible & -0.0013 & \textbf{0.0001}\\ 
Informative & \textbf{0.0032} & 0.0009\\ 
Inquisitive & \textbf{0.0095} & -0.003\\ \hline
\end{tabular}
\caption{FED scores of dialogues generated with and without CoT for DialogSum few-shot}
\label{tab:ablation_cot_dialogsum_fed}
\end{table}

\begin{table}[H]
\centering
\begin{tabular}{lcc}
\hline
\textbf{Criteria} & \textbf{Without CoT} & \textbf{With CoT} \\ \hline
Coherence & 0.0052 & \textbf{0.0284} \\ 
Diversity & 0.0120 & \textbf{0.0303} \\ 
Flexibility & 0.0074 & \textbf{0.0215}\\ 
Understandability & 0.0056 & \textbf{0.019}\\ 
Inquistiveness & 0.0162 & \textbf{0.0362}\\ 
Consistency & 0.0091 & \textbf{0.0366}\\ 
Informativeness & 0.0099 & \textbf{0.0168}\\ 
Likeability & 0.0029 & \textbf{0.0209} \\ 
Depth & 0.0062 & \textbf{0.0115} \\ 
Error Recovery & 0.0142 & \textbf{0.0242}\\ \hline
\end{tabular}
\caption{GPTScore of dialogues generated with and without CoT for DialogSum few-shot}
\label{tab:ablation_cot_dialogsum_gptscore}
\end{table}

\begin{table}[H]
\centering
\begin{tabular}{lcc}
\hline
\textbf{Criteria} & \textbf{Without CoT} & \textbf{With CoT} \\ \hline
Coherent & \textbf{0.9667} & 0.9125 \\ 
Error Recovery & \textbf{0.9577} & 0.9051 \\ 
Consistent & \textbf{0.9621} & 0.9163 \\ 
Diverse & \textbf{0.9586} & 0.9124 \\ 
Depth & \textbf{0.9589} & 0.9094 \\ 
Likeable & \textbf{0.0059} & -0.0003 \\ 
Understand & \textbf{0.9503} & 0.8978 \\ 
Flexible & \textbf{-0.0022} & -0.0023 \\ 
Informative & \textbf{0.0110} & 0.0035 \\ 
Inquisitive & \textbf{0.0030} & -0.0037 \\ \hline
\end{tabular}
\caption{FED scores of dialogues generated with and without CoT for SAMSum few-shot}
\label{tab:ablation_cot_samsum_fed}
\end{table}

\begin{table}[h]
\centering
\begin{tabular}{lcc}
\hline
\textbf{Criteria} & \textbf{Without CoT} & \textbf{With CoT} \\ \hline
Coherence & 0.0076 & \textbf{0.0324} \\
Diversity & 0.0172 & \textbf{0.0372} \\ 
Flexibility & 0.0116 & \textbf{0.0259}\\ 
Understandability & 0.0104 & \textbf{0.0272}\\ 
Inquistiveness & 0.0201 & \textbf{0.0389}\\ 
Consistency & 0.0149 & \textbf{0.0415}\\ 
Informativeness & 0.0148 & \textbf{0.0200}\\ 
Likeability & 0.0041 & \textbf{0.023} \\ 
Depth & 0.0078 & \textbf{0.0124} \\ 
Error Recovery & 0.0163 & \textbf{0.0289}\\ \hline
\end{tabular}
\caption{GPTScore of dialogues generated with and without CoT for SAMSum few-shot}
\label{tab:ablation_cot_samsum_gptscore}
\end{table}

\section{Ablation Studies on Subtopics and Personas}
\label{app:ablation_subtopics_personas}

To further validate the effectiveness of the DiaSynth framework, we conducted ablation studies 
by evaluating the impact of removing sub-topics and personas from the data generation pipeline. 
The goal was to assess their contribution to the quality of the generated dialogues. 

For these experiments, we generated approximately 960-1000 dialogues using Phi-3 and compared 
three settings:
\begin{itemize}
    \item \textbf{subtopics}: Removing sub-topics while keeping personas.
    \item \textbf{personas}: Removing personas while keeping sub-topics.
    \item \textbf{diasynth}: The full DiaSynth-generated data with both personas and sub-topics.
\end{itemize}

\begin{table}[h]
    \centering
    \renewcommand{\arraystretch}{1.1} % Adjust row spacing
    \setlength{\tabcolsep}{2pt} % Reduce column spacing
    \begin{tabular}{l p{1.5cm} p{1.5cm} p{1.5cm}}
        \toprule
        \textbf{Metric} & \textbf{subtopics} & \textbf{personas} & \textbf{diasynth} \\
        \midrule
        Coherent & 0.9252 & \textbf{0.9584} & 0.9536 \\
        Error Recovery & 0.9022 & 0.9414 & \textbf{0.944} \\
        Consistent & 0.9095 & 0.9512 & \textbf{0.954} \\
        Diverse & 0.9139 & 0.9512 & \textbf{0.9534} \\
        Depth & 0.9193 & \textbf{0.9533} & 0.9521 \\
        Likeable & 0.0069 & \textbf{0.007} & -0.0005 \\
        Understandable & 0.8918 & 0.9339 & \textbf{0.9353} \\
        Flexible & -0.0038 & -0.0042 & \textbf{0} \\
        Informative & \textbf{0.0096} & 0.0072 & 0.009 \\
        Inquisitive & \textbf{0.0129} & 0.0063 & -0.0033 \\
        \bottomrule
    \end{tabular}
    \caption{FED scores with DialogSum as base}
    \label{tab:fed_dialogsum_ablation}
\end{table}

\begin{table}[h]
    \centering
    \renewcommand{\arraystretch}{1.1} % Adjust row spacing
    \setlength{\tabcolsep}{2pt} % Reduce column spacing
    \begin{tabular}{l p{1.5cm} p{1.5cm} p{1.5cm}}
        \toprule
        \textbf{Metric} & \textbf{subtopics} & \textbf{personas} & \textbf{diasynth} \\
        \midrule
        Coherence & 0.0118 & 0.0098 & \textbf{0.0286} \\
        Diversity & 0.0246 & 0.0198 & \textbf{0.031} \\
        Flexibility & \textbf{0.0239} & 0.0121 & 0.0193 \\
        Understandable & 0.0137 & 0.0121 & \textbf{0.0363} \\
        Inquisitive & 0.0358 & 0.0308 & \textbf{0.0363} \\
        Consistent & 0.0175 & 0.0132 & \textbf{0.0369} \\
        Informative & \textbf{0.0242} & 0.0155 & 0.0201 \\
        Likeability & 0.0083 & 0.0076 & \textbf{0.0213} \\
        Depth & \textbf{0.0216} & 0.0181 & 0.0117 \\
        Error Recovery & 0.0337 & 0.0249 & \textbf{0.0342} \\
        \bottomrule
    \end{tabular}
    \caption{GPTScore with DialogSum as base}
    \label{tab:gptscore_dialogsum_ablation}
\end{table}

\begin{table}[h]
    \centering
    \renewcommand{\arraystretch}{1.1} % Adjust row spacing
    \setlength{\tabcolsep}{2pt} % Reduce column spacing
    \begin{tabular}{l p{1.5cm} p{1.5cm} p{1.5cm}}
        \toprule
        \textbf{Metric} & \textbf{subtopics} & \textbf{personas} & \textbf{diasynth} \\
        \midrule
        Coherence & 0.0106 & 0.0109 & \textbf{0.0325} \\
        Diversity & 0.0224 & 0.0191 & \textbf{0.0372} \\
        Flexibility & 0.0235 & 0.0141 & \textbf{0.0395} \\
        Understandable & 0.0159 & 0.0252 & \textbf{0.0415} \\
        Inquisitive & 0.0334 & 0.0137 & \textbf{0.0415} \\
        Consistent & 0.0192 & \textbf{0.0415} & 0.037 \\
        Informative & \textbf{0.0228} & 0.0155 & 0.021 \\
        Likeability & 0.0128 & 0.0075 & \textbf{0.0232} \\
        Depth & \textbf{0.0187} & 0.0158 & 0.0126 \\
        Error Recovery & \textbf{0.0323} & 0.0213 & 0.029 \\
        \bottomrule
    \end{tabular}
    \caption{FED scores with SAMSum as base}
    \label{tab:fed_samsum_ablation}
\end{table}

\begin{table}[h]
    \centering
    \renewcommand{\arraystretch}{1.1} % Adjust row spacing
    \setlength{\tabcolsep}{2pt} % Reduce column spacing
    \begin{tabular}{l p{1.5cm} p{1.5cm} p{1.5cm}}
        \toprule
        \textbf{Metric} & \textbf{subtopics} & \textbf{personas} & \textbf{diasynth} \\
        \midrule
        Coherence & 0.0106 & 0.0109 & \textbf{0.0325} \\
        Diversity & 0.0224 & 0.0191 & \textbf{0.0372} \\
        Flexibility & 0.0235 & 0.0141 & \textbf{0.0395} \\
        Understandable & 0.0159 & 0.0252 & \textbf{0.0415} \\
        Inquisitive & 0.0334 & 0.0137 & \textbf{0.0415} \\
        Consistent & 0.0192 & \textbf{0.0415} & 0.037 \\
        Informative & \textbf{0.0228} & 0.0155 & 0.021 \\
        Likeability & 0.0128 & 0.0075 & \textbf{0.0232} \\
        Depth & \textbf{0.0187} & 0.0158 & 0.0126 \\
        Error Recovery & \textbf{0.0323} & 0.0213 & 0.029 \\
        \bottomrule
    \end{tabular}
    \caption{GPTScore with SAMSum as base}
    \label{tab:gptscore_samsum_ablation}
\end{table}

The ablation studies presented in Tables \ref{tab:fed_dialogsum_ablation}, \ref{tab:gptscore_dialogsum_ablation}, 
\ref{tab:fed_samsum_ablation} and \ref{tab:gptscore_samsum_ablation} demonstrate that the inclusion of both personas and sub-topics significantly 
enhances the quality of generated dialogues across FED and GPTScore metrics. For both DialogSum 
and SAMSum few-shot examples, dialogues generated with the full DiaSynth framework, incorporating both personas 
and sub-topics, achieved the highest scores in coherence, diversity, and consistency. This indicates 
that structured dialogue generation benefits from incorporating diverse sub-topics while maintaining 
persona-driven consistency.

An interesting observation arises in Table~\ref{tab:fed_samsum_ablation}, where for the 
SAMSum dataset, the best-performing configuration involved using only sub-topics without personas. 
This deviation can likely be attributed to the more informal nature of SAMSum dialogues, where 
structured personas introduce a formal communication style that does not align well with the dataset. 
In contrast, in more structured datasets like DialogSum, the addition of personas provides clear 
improvements, ensuring dialogue coherence and natural flow.

Moreover, the largest improvements in quality are seen in coherence, error recovery, and 
understandability, particularly when both sub-topics and personas are included. While sub-topics 
alone contribute significantly to improving diversity and depth, their combination with 
personas enhances overall dialogue quality. This suggests that dataset characteristics play a 
crucial role in determining the effectiveness of persona modeling, highlighting the need for 
adaptive strategies in synthetic dialogue generation. Ultimately, these findings reinforce that 
DiaSynth-generated dialogues are robust and adaptable, providing high-quality synthetic data across 
both structured and informal conversational settings.

\section{Additional Methodological Details}
\label{app:method_details}

\subsection{Prompts for Dialogue and Summarization Generation}
\label{sec:prompts}

The prompts used for dialogue generation were carefully designed to guide the model in producing contextually rich and persona-driven conversations. The structure ensures that dialogues exhibit natural human-like characteristics, considering aspects such as speaker familiarity, emotional states, and formality levels.

For dialogue generation, the following detailed prompt was used:

% \begin{lstlisting}[language=Python, caption={Prompt used for dialogue generation}, label={lst:dialogue_prompt}]
\begin{quote}
You are an expert dialog generator. 
The following are examples of real-life dialogues.

example 1

$ "dialogue" - "{dialogue_1}" $

example 2

$ "dialogue" - "{dialogue_2}" $

Use the examples as references and generate a dialogue between people with the following personas:

$ persona 1 - "{persona_1}" $

$ persona 2 - "{persona_2}" $

Characteristics of the dialogue - to be understood before generating the dialogue:
{characteristics}

- Assume the values for the characteristics and provide an explanation for choosing those values.

- These characteristics should be well understood as they implicitly affect and guide the conversation.

Chain of Thought Reasoning:

- Before generating the dialogue, reason about the values for each characteristic listed above.

- Explain in detail how each characteristic will influence a hypothetical dialogue between persons with those characteristics and personas.

- Ensure that the reasoning considers the interactions between different characteristics (e.g., how familiarity and emotional state might interact).

- This reasoning and explanation must be included between <cot> and </cot> tags. Do not skip this step.

- The dialogue generated should be based on the explanation provided.
\end{quote}

This prompt ensures that the model generates structured and realistic dialogues by incorporating a 
reasoning step before dialogue generation. The inclusion of CoT reasoning forces the model to explicitly 
consider multiple conversational attributes, leading to more coherent and contextually appropriate responses.

This structured prompting approach enables the generation of high-quality dialogues. The complete list of 
prompts, including variations for different tasks, is available in our public code repository.

\subsection{Seed Topic Selection}
The seed topics were selected to ensure diversity across conversational scenarios. The topics 
used in our experiments are as follows:

\begin{quote}
Remote Work, Book Recommendations, Fitness Routines, Travel Destinations, Career Development, Movie Reviews, Video Games, Pets, Stock Market, Fashion Trends, Online Education, Mental Wellness, Climate Change, Sports, Artificial Intelligence, Food.
\end{quote}

These topics were chosen from a mix of real-world discussion trends and common conversational themes, ensuring broad coverage and relevance for synthetic dialogue generation.

\subsection{Hyperparameters for Generation}
\label{sec:hyperparams}
The hyperparameter settings for different stages of generation are reported below:

\paragraph{Subtopic Generation:} Temperature: 0.1 and max tokens: 2048
\paragraph{Persona Generation:} Temperature: 0.1 and max tokens: 2048
\paragraph{Dialogue Generation:} Temperature: 0.2 and max tokens: 4096

\section{Additional Downstream Task: Response Generation}
\label{app:response_generation}

\begin{table*}[h]
    \centering
    \begin{tabular}{lcccc}
        \toprule
        \textbf{Model} & \textbf{Before Fine-tuning} & \textbf{In-domain Data} & \textbf{Llama3} & \textbf{GPT-4o} \\
        \midrule
        t5-base  & 0.4003  & 0.6870  & 0.6572  & 0.6612 \\
        bart-base  & 0.5681  & 0.6875  & 0.6721  & 0.6630 \\
        \bottomrule
    \end{tabular}
    \caption{BERTScore evaluation of response generation models fine-tuned on synthetic and in-domain datasets.}
    \label{tab:response_gen}
\end{table*}

To further evaluate the utility of DiaSynth-generated data, we conducted an additional 
downstream task: \textbf{response generation}. This task was included to validate the 
effectiveness of our synthetic data beyond summarization. For these experiments, we selected \textbf{Llama3} 
and \textbf{GPT-4o} as base datasets since they demonstrated 
superior performance across both quality and summarization metrics. We employed \textbf{BERTScore} as our
evaluation metric due to its effectiveness in measuring the similarity between generated and 
reference responses. The results are presented in Table~\ref{tab:response_gen}.

Both models show a significant improvement in BERTScore after fine-tuning on DiaSynth-generated data 
compared to the pre-trained baseline, highlighting its effectiveness in enhancing model performance. 
Additionally, models fine-tuned on DiaSynth data achieve scores that are 
\textbf{close to those fine-tuned on in-domain data}. For instance, \textbf{t5-base} achieves a 
BERTScore of \textbf{0.6572} on Llama3-generated data and \textbf{0.6612} on GPT-4o-generated data, 
compared to \textbf{0.6870} for in-domain fine-tuning. These results indicate that DiaSynth-generated 
data serves as a viable alternative for fine-tuning response generation models, performing comparably to 
in-domain data, even in low-resource scenarios.

\end{document}